\documentclass[runningheads]{llncs}
\usepackage{graphicx}
\usepackage{amsmath}
\usepackage{amssymb}
\usepackage{xcolor}
\usepackage{url}
\usepackage{tabularx}

\newcommand{\model}{{GATv2-TCN}}
\newcolumntype{Y}{>{\centering\arraybackslash}X}

\begin{document}
\title{Who You Play Affects How You Play: Predicting Sports Performance Using Graph Attention Networks With Temporal Convolution} %
\author{Rui Luo\inst{1}\orcidID{0000-0003-0711-8039} \and
Vikram Krishnamurthy\inst{1}\orcidID{0000-0002-4170-6056}}
\authorrunning{R. Luo and V. Krishnamurthy}
\titlerunning{Sports Performance Prediction Using \model{}}
\institute{ %
Cornell University
\email{\{rl828, vikramk\}@cornell.edu}\\
}
\maketitle              %
\begin{abstract}
This study presents a novel deep learning method, called \model{}, for predicting player performance in sports. To construct a dynamic player interaction graph, we leverage player statistics and their interactions during gameplay. We use a graph attention network to capture the attention that each player pays to each other, allowing for more accurate modeling of the dynamic player interactions. To handle the multivariate player statistics time series, we incorporate a temporal convolution layer, which provides the model with temporal predictive power. We evaluate the performance of our model using real-world sports data, demonstrating its effectiveness in predicting player performance. Furthermore, we explore the potential use of our model in a sports betting context, providing insights into profitable strategies that leverage our predictive power. The proposed method has the potential to advance the state-of-the-art in player performance prediction and to provide valuable insights for sports analytics and betting industries.

\end{abstract}
\section{Introduction}
Due to the unpredictable nature of sports and the seemingly countless variables that could have an effect, sports prediction is a topic of significant interest. The majority of methods currently in use forecast a player's performance in the current game using her statistics from past games. We suggest that predicting player performance involves a complicated system with a web of interdependencies between teammates, rivals, and even coaches, all of which may influence a player's performance. 

This study aims to predict player performance by taking into account their interactions. To this end, we construct a dynamic player interaction graph. 
When updating a player's feature using other players in interaction, we employ a graph attention network to calculate the attention that each player pays to each other. 
We propose a model called \model{} that combines the graph attention network with a temporal convolution layer and outperforms existing models that solely rely on player historical performance. We also conduct experiments to demonstrate how to profit on a sports betting website using the predictive power of this model.

An open-sourced sports dataset with both dynamic graph structure and temporal node attributes is one of the study's main contributions. We anticipate that this dataset will be useful for future temporal geometric machine learning research.

\section{Related Work}
The majority of studies on sports prediction to date have focused on forecasting game outcomes rather than individual player performance. This study uses a dynamic player interaction graph to reflect players' interactions. We transform the task of predicting player performance into one involving temporal geometric learning. The following provides a summary of the literature on temporal geometric deep learning, player performance prediction, and game result prediction.

\subsection{Game Result Prediction}
One of the earliest studies to utilize artificial neural networks (ANNs) to predict National Basketball Association (NBA) game results was Loeffelholz et al. \cite{loeffelholz2009predicting}'s work. 620 matches were used for training and testing, whereas 30 matches were used as the validation set.  
Cao \cite{cao2012sports} examined several models, including logistic regression, naive Bayes, support vector machine, and ANN, based on data from over 4000 NBA matches over the course of six seasons (2005-2010). 
To identify the best collection of features, Thabtah et al. \cite{thabtah2019nba} trained naive Bayes models and ANNs on 430 NBA Finals matchups. The most significant variable discovered to affect match results was defensive rebounds. 
A Gamma process-based in-play prediction model for the NBA scoring process was presented by Song and Shi \cite{song2020gamma}. The model also accounted for the dependence between the home team and the visiting team. 

For a review on predicting sports results using machine learning, we direct readers to \cite{bunker2022application}. The main conclusion is that choosing and engineering a suitable feature set seems to be more crucial than having a lot of data. %

\subsection{Player Performance Prediction}
Since 2015, a player prediction method named \textit{Career-Arc Regression Model Estimator with Local Optimization} (CARMELO)\footnote{\url{https://projects.fivethirtyeight.com/carmelo/}}, which incorporates player performance forecasting into game prediction, has been a part of the well-known FiveThirtyEight model to predict NBA games\footnote{\url{https://fivethirtyeight.com/methodology/how-our-nba-predictions-work/}}. CARMELO compares current players to those from the past who had statistically similar profiles to the same age and produces a probabilistic projection for the remainder of the player's career. On a similar note, Ishi and Patil \cite{ishi2021study} used features such as previous record and recent performance to find the strengths and weaknesses of the players and decide the optimal cricket team formation. 
Oytun et al. \cite{oytun2020performance} investigated machine learning models including ANNs and Long short-term memory (LSTM) neural networks to predict female handball players performance and to identify the major influential parameters.

Other studies have looked at how a player affects the outcome of the team. 
Berri \cite{berri1999most} used an econometric model to relate NBA player data to team victories. The marginal impact obtained from the model reflects the efficiency of a single player in a game. 
After adjusting for the other players on the court, Deshpande and Jensen \cite{deshpande2016estimating} developed a Bayesian linear regression model to estimate an individual player's impact on their teams' chances of winning. The model result reveals players whose significant contribution is underrepresented by existing metrics. %
Additionally, computational models that are tailored for various sorts of data have been created to predict player performance. Oved et al. \cite{oved2020predicting} studied the pre-game and post-game media interviews and measured the correlation between the interview transcripts and player performance. Their text-based models outperform baselines trained only on performance metrics.

\subsection{Temporal Geometric Deep Learning}
The interaction and influence between players (both teammates and opponents) in a match has not received much research attention. Our research aims to bridge the knowledge gaps in this area. To solve this problem, we resort to temporal geometric deep learning method. 

Lea et al. \cite{lea2017temporal} proposed a temporal convolution network (TCN) for video-based action segmentation. TCN unified the local spatial-temporal feature extraction and high-level temporal pattern detection with an encoder-decoder architecture where a hierarchy of temporal convolutions is applied. 
Guo et al. \cite{guo2019attention} presented an attention based spatial-temporal graph convolution model (ASTGCN) for highway traffic forecasting. The model combines the spatial-temporal attention mechanism and the spatial-temporal convolution, where the Chebyshev graph convolution (ChebyConv) is applied in the spatial dimension. 
Wu et al. \cite{wu2020connecting} proposed a graph neural network model (MTGNN) for multivariate time series forecasting. They considered multivariate time series variables as nodes in graphs and used a graph convolution and a temporal convolution to capture the spatial and temporal dependencies respectively. 
Dai et al. \cite{dai2020hybrid} proposed a Hybrid Spatio-Temporal Graph Convolutional Network (H-STGCN) for traffic forecasting. They used a transformer structure to and graph convolution network (GCN) to capture the spatial dependency. 
Bi et al. \cite{bi2022pangu} presented a 3D Earth-specific transformer (3DEST) extended from the 2D vision transformer block \cite{dosovitskiy2020image} that suits 3D weather states. The transformer uses a specific positional bias to align with the earth's geometry and capture the uneven spatial distribution of earth's sphere. 

Many of the aforementioned approaches are devoted to the traffic forecasting problem, which is similar to the player performance prediction problem in that not only can the traffic network (player network) be explicitly defined without inferring as a hidden variable, the traffic (performance) of one road (player) has weekly or daily pattern (game momentum \cite{arkes2011finally}; or the "hot hand" effect \cite{bar2006twenty}). And it depends on the traffic (performance) of other roads (players) that intersect with it.

\section{Dataset Description}\label{sec:data}
The availability of extensive datasets like the Open Graph Benchmark (OGB) \cite{hu2020open} has aided recent advancements in geometric deep learning, particularly, graph neural networks (GNNs). Many existing GNNs, such as graph convolution network (GCN) \cite{kipf2016semi}, Chebyshev polynomial approximated spectral graph convolution network (ChebyConv) \cite{defferrard2016convolutional}, and graph attention network (GAT) \cite{velickovic2017graph}, however, are not directly applicable to temporal geometric deep learning challenges because they are constrained by either a fixed topology or static node and edge features. 

Rozemberczki et al. \cite{rozemberczki2021pytorch} presented an open-source geometric deep learning library\footnote{\url{https://pytorch-geometric-temporal.readthedocs.io/en/latest/index.html}} which includes several benchmark datasets for tasks such as social media behavior prediction, epidemiological forecasts, and energy production. However, there is no sports dataset in this field. We aim to fill this gap by providing a sports dataset with both dynamic graph structure and temporal node features. 

\begin{table}
\caption{NBA player statistics used in the dataset. It covers 7 basic statistics, 3 tracking statistics, and 3 advanced statistics.}\label{table:stats}
\begin{center}
\scriptsize
\begin{tabularx}{\textwidth}{|l|X|}
\hline
Abbreviation &  Definition \\
\hline
PTS (Points) &  the number of points scored \\
AST (Assists) & the number of passes that lead directly to a made basket\\
REB (Rebounds) & the number of events when a player recovers the ball after a missed shot\\
TO (Turnovers) & a turnover occurs when the team on offense loses the ball to the defense\\
STL (Steals) & the number of times a defensive team takes the ball from a player on offense, causing a turnover\\
BLK (Blocks) & the number of evetns when an offensive player attempts a shot, and the defense player tips the ball, blocking their chance to score \\
PLUS\_MINUS (Plus-Minus) & the point differential when a team is on the floor \\
\hline
TCHS (Touches) & the number of times a player or team touches and posseses the ball during the game\\
PASS (Passes) & the number of passes made to or received from the given player or team\\
DIST (Distance Miles) & distance run by a player or team measured in miles\\
\hline
PACE (Pace) & the number of possessions of a player per 48 minutes \\
USG\_PCT (Usage Percentage) & the percentage of team plays used by a player when they are on the floor \\
TS\_PCT (True Shooting Percentage) & a shooting percentage that factors in the value of three-point field goals and free throws in addition to conventional two-point field goals\\
\hline
\end{tabularx}
\end{center}
\end{table}

\subsection{Data Collection}
\subsubsection{Player Statistics}
The dataset covers the 2022-23 NBA regular season (2022-10-18 to 2023-01-20) which contains 691 games in 92 game days. There are 582 active players among the 30 teams. Besides 7 basic statistics, we collected 3 tracking statistics, and 3 advanced statistics. We use tracking statistics to more accurately reflect players' movements on the court, and advanced statistics to more properly represent a player's effectiveness and contribution to the game. Together, these two types of data give us a better understanding of factors that are not visible on the scoreboard. Table~\ref{table:stats} defines the 13 player statistics included in the dataset. 

\subsubsection{Match Information}
We use the match information to construct player interaction graphs (Section \ref{subsec:graph construction}). On each game day, we obtain the roster for each game. We construct a complete graph that contains all players from both teams who played for at least 10 minutes. This accounts for the impacts of bench players on the game. 

All the data is freely accessed through the APIs of NBA.com using a Python API Client Package\footnote{\url{https://github.com/swar/nba_api}}. We use four endpoints, \textit{boxscoretraditionalv2}, \textit{boxscoreplayertrackv2}, \textit{boxscoreadvancedv2}, and \textit{leaguegamefinder} to fetch the basic statistics, the tracking statistics, and the advanced statistics, and the match information respectively. We use another two endpoints, \textit{} for getting the team roster and 

\subsection{Exploratory Data Analysis}
\begin{figure}
\includegraphics[width=1\textwidth]{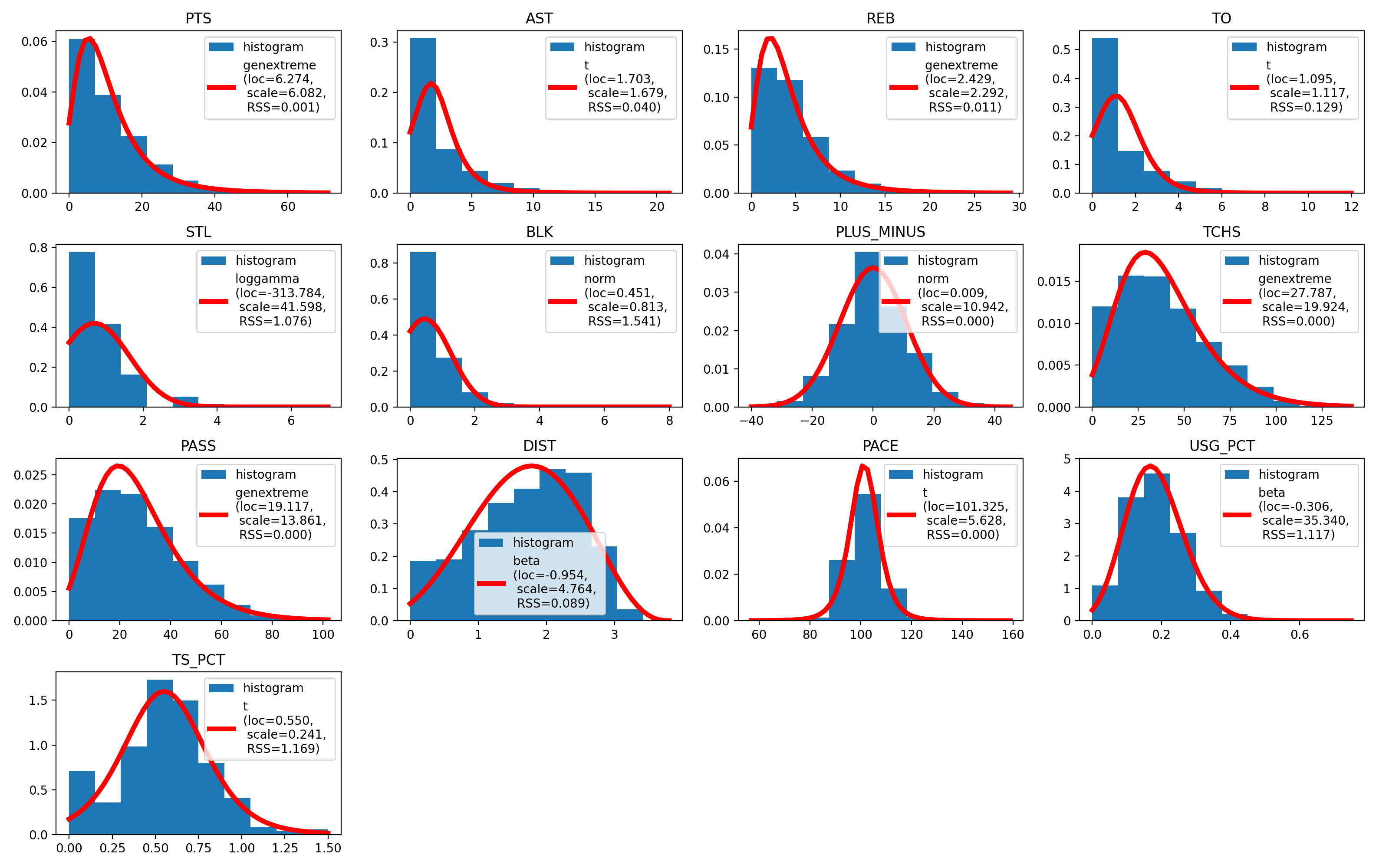}
\caption{The histogram and best-fit distribution of the 13 player statistics collected in the dataset. The legend shows the location parameter, scale parameter, and the residual sum of squares (RSS) of the best-fit distribution.} \label{fig:data distribution}
\end{figure}

We fitted each player statistic during the period using various distributions including normal distribution (\textit{norm}), $t$-distribution (\textit{t}), exponential distribution (\textit{exp}), beta distribution (\textit{beta}), gamma distribution (\textit{gamma}), and generalized extreme value distribution (\textit{genxetreme}). We use the residual sum of squares (RSS) to decide the distribution that best fits the statistic. Figure \ref{fig:data distribution} shows the best-fit distribution against the histogram of each statistic. The legend shows the location parameter, scale parameter, and the RSS.

\section{Methodology: Graph Attention Network with Temporal Convolution Layer}
In this section, the task of predicting player performance is formulated as a multivariate time series forecasting problem. The dynamic player interaction graph is constructed on each game day based on the roster information. 
We present the proposed graph attention based temporal convolution model (\model{}), which is comprised of two major components: the graph attention network (GATv2) \cite{GATv2}; and the temporal convolution layer (TCN) \cite{guo2019attention}. 

\subsection{Player Interaction Graph Construction}\label{subsec:graph construction}
Based on the the roster information from each game, we construct a dynamic player interaction graph. 
We built the player graph explicitly, as opposed to inferring the latent graph structure from multivariate time series, such as the graph learning layer in \cite{wu2020connecting}. 

Denote $G^{(1)}, \cdots, G^{(T)}$ as a sequence of graph snapshots. $G^{(t)} = (V^{(t)}, E^{(t)}), 1 \leq t \leq T$ represents the graph observed at discrete time $t$. In particular, $G^{(t)}$ corresponds to the games that take place on a single game day or over a set period of time. 
$V^{(t)}$ is the set of $n$ players involved in games at time $t$. We restrict that $V^{(t)} = V =\{v_i\}_{i=1, \cdots, n}, 1 \leq t \leq T$ so that all the graph snapshots have the same set of players, i.e. the league roster is fixed over the course considered. $E=\{\{ v_i, v_j \}\}$ is the set of \textit{undirected} edges\footnote{We omit edge direction for expositional simplicity and because most sports demand that players handle offense and defense on both sides. The approach is generalizable to directed networks for sports like baseball and American football where players are further divided into distinct offense/defense roles.} between players. 
We add the edges as follows: if there is a game between two teams $V_A, V_B \subset V, V_A \cap V_B = \varnothing$ at time $t$, then we form an undirected edge between any two players that played for at least 10 minutes from the two teams, i.e., we create a complete graph among the players $V_A \sqcup V_B$. In other words, $G^{(t)}$ is a cluster graph \cite{shamir2004cluster}. 
The adjacency matrix $A^{(t)}$ of $G^{(t)}$ is a $n \times n$ symmetric square matrix where its entry
\begin{equation}
    a^{(t)}_{ij} = 
    \begin{cases}
        1 & \textrm{if } \{v_i, v_j\} \in E^{(t)} \\
        0 & \textrm{otherwise.}
    \end{cases}
\end{equation}
The degree matrix $D^{(t)}$ is defined as a diagonal matrix with entries $d^{(t)}_{ii} = \sum_{j=1}^n a^{(t)}_{ij}$. The graph Laplacian is defined as $L^{(t)} = D^{(t)} - A^{(t)}$. Next, we demonstrate that the expressivity of spectral graph neural networks (e.g., GCN and ChebyConv) based on Laplacian spectrum is limited in the case of player interaction graphs. 

Suppose that at time $t$, there are $p$ games where game $m \in \{1, \cdots, p\}$ involves $k_m$  players. Note that $L^{(t)}$ is a block diagonal matrix 
\begin{equation}
    L^{(t)} = \begin{bmatrix}
        L^{(t)}_1 & 0 & \cdots & 0 & 0\\
        0 & L^{(t)}_2 & \cdots & 0 & 0\\
        \vdots & \vdots & \ddots & \vdots & \vdots \\
        0 & 0 & \cdots & L^{(t)}_p & 0 \\
        0 & 0 & 0 & 0 & 0 \\
    \end{bmatrix},
\end{equation}
where each block $L^{(t)}_m$ is the graph Laplacian for a complete graph with $k_m$ nodes, 
\begin{equation}
    L^{(t)}_m = k_m I_{k_m} - J_{k_m},
\end{equation}
where $J_{k_m} = \boldsymbol{1}\boldsymbol{1}^{\top}$ is the all-ones matrix. 

The eigenvalues of $L^{(t)}$ are $0^{n-\sum_{m=1}^{p} k_m + p}, k_1^{k_1 - 1}, \cdots, k_p^{k_p - 1}$. All the eigenvalues are repeated. In addition, the eigenvalue distribution is sparse because $k_1, \cdots, k_p$ are usually close to each other. According to Theorem 4.5 in \cite{wang2022powerful}, player interaction graphs have a high order of automorphism; And spectral GNNs fail to produce all one-dimensional predictions in the absence of nonlinearity, implying the use of other graph neural networks.

In Section \ref{sec:experiment}, we show that a model based on spectral GNN (ASTGCN) underperforms the proposed \model{} based on graph attention. In the following Section \ref{subsec:GAT}, we discuss the implementation of GATv2 on player interaction graphs.

\subsection{Learning Player Attention with GATv2}\label{subsec:GAT}
Given the constructed player interaction graphs, we demonstrate how GATv2 unveil the relationship and influence between players. We associate each player $v_i \in V$ with a feature vector $f_i^{(t)}$ at time $t$. Each dimension may represent one statistic of the player.  

To reflect the effects of team and position on the attention one node gets from another, we also build two linear transformations to learn the team embeddings and position embeddings, and concatenate the player's feature vector with the embedding vectors $g_i^{(t)} = [f_i^{(t)} \| team_i \| pos_i] \in \mathbb{R}^{d}$. 

First, at each time $t$, a weight matrix $W \in \mathbb{R}^{d' \times d}$ is applied to every player to transform the feature vector linearly a new set of player representations, $W f_i^{(t)}$. Then, a self-attention is performed on the transformed player representations which computes a score representing the importance of the features of a neighboring player $j$ to the player $i$:
\begin{equation}\label{eq:scoring}
    e(g_i^{(t)}, g_j^{(t)}) = a^{\top} \textrm{LeakyReLU}(W[g_i^{(t)} \| g_j^{(t)}])
\end{equation}
where $a \in \mathbb{R}^{2d'}$ is a learnable weight vector applied after the nonlinearity ($\textrm{LeakyReLU}$) and $\|$ is the concatenation operation. Note that we view undirected edges as bidirectional, i.e., each edge $\{v_i, v_j \}$ can be represented as a pair of tuples $(v_i, v_j)$ and $(v_j, v_i)$. This enables improved expressivity as the influence of $v_i$ to $v_j$ may differ from that of $v_j$ to $v_i$.

\begin{figure}
\includegraphics[width=1\textwidth]{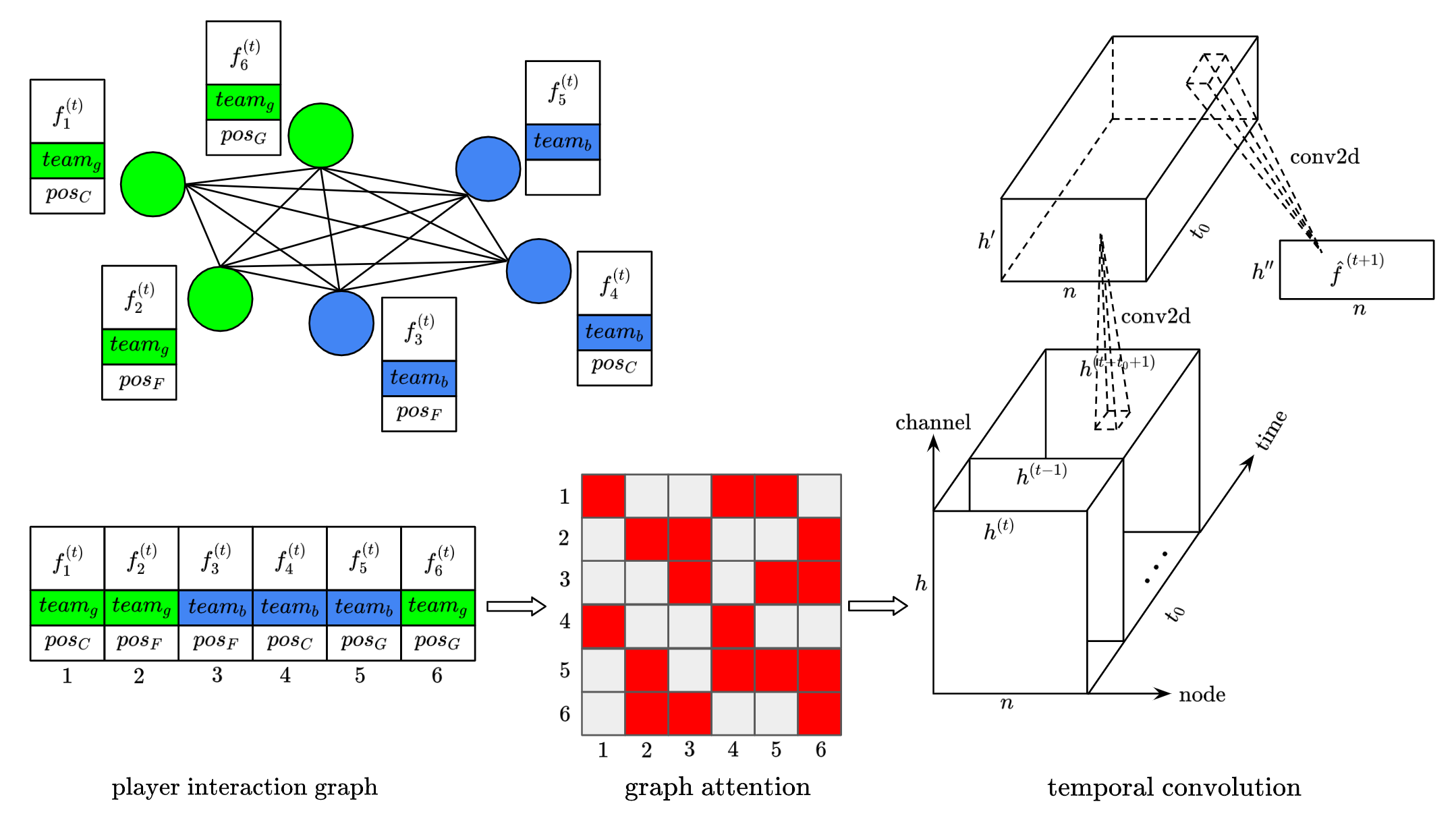}
\caption{The diagram shows the structure of the proposed \model{}. 
The player interaction graph on the top left (defined in Section \ref{subsec:graph construction}) is a complete graph with 6 nodes where the node color indicates team. The feature vector is concatenated with the team embedding and position embedding $[f_i^{(t)}\|team_i\|pos_i], i=1,\cdots,6$. The node representation $h(t)$ (\ref{eq:GAT output}) is obtained by propagating the concatenated vectors through the edges based on weights calculated using the graph attention (\ref{eq:attention}). A temporal convolution layer that fuses $h^{(t)}, \cdots, h^{(t-t_0+1)}$ is applied to produce the predicted node features $\hat{f}^{(t+1)}$ at time $t+1$.} \label{fig:diagram}
\end{figure}

The scores computed in (\ref{eq:scoring}) are then normalized across all neighbors of player $l\in \mathcal{N}_i$ which lead to the attention function:
\begin{equation}\label{eq:attention}
    \alpha_{ij}^{(t)} = \textrm{softmax}(e(g_i^{(t)}, g_j^{(t)})) = \frac{\textrm{exp}(e(g_i^{(t)}, g_j^{(t)}))}{\sum_{l \in \mathcal{N}_i} \textrm{exp}(e(g_i^{(t)}, g_l^{(t)}))}
\end{equation}
Finally, GATv2 updates the player representation by weighting the attention function and propagating the prior representation:
\begin{equation}\label{eq:GAT output}
    h_i^{(t)} = \sigma(\sum_{j \in \mathcal{N}_i} \alpha_{ij}^{(t)} W g_i^{(t)}) \in \mathbb{R}^{h}
\end{equation}
where $\sigma$ represents a nonlinearity function and $h$ is the dimension of the output. Similar to multi-head attention in Transformers, we follow the design in \cite{GAT,GATv2} to use $H$ separate attention heads and concatenate their output representations.

Note that in the original GAT model \cite{GAT}, the scoring function is 
\begin{equation}
    e(g_i^{(t)}, g_j^{(t)}) =  \textrm{LeakyReLU}(a^{\top} W[g_i^{(t)} \| g_j^{(t)}]), 
\end{equation}
as opposed to (\ref{eq:scoring}) where the linear transformation ($a$) is applied after the nonlinearity ($\textrm{LeakyReLU}$). In other words, the two linear layers ($W$ and $a$) are applied immediately after one another, which reduces the expressivity of the self-attention and fails to compute the dynamic attention \cite{GATv2}. %

\subsection{Combining Graph Attention with Temporal Convolution}
The objective of the player performance prediction is to predict $f^{(t)}$ based on the sequence of $t_0$ previous features $f^{(t-t_0)}, \cdots, f^{(t-1)}$. Due to variances in weekdays and weekends, holidays, and other events, game days may not be evenly spread in time. The amount of games a player has played as of a certain time $t$ may vary, and they may have missing data for times when their team does not play. 
We use the forward filling, a method that is frequently used in time series forecasting, to fill in the missing values with previous data.

Although the player representation (\ref{eq:GAT output}) produced by GATv2 reflects the chemistry among players at a given time, it does not capture the evolving dynamics of the chemistry or the player's individual development. To help the model attain its prediction power, we further implement a convolution layer in the temporal dimension which updates the player representation by merging the information from the neighboring time slice \cite{guo2019attention}. Denote $H^{(t)} = [h^{(t-t_0 + 1)}, \cdots, h^{(t)}] \in \mathbb{R}^{n\times h \times t_0}$, the output of temporal convolution layer is 
\begin{equation}
    Y^{(t)} = \Phi * \textrm{ReLU}(H^{(t)})
\end{equation}
where $\Phi$ is the kernel of the temporal convolution and the activation function is ReLU. Finally, a fully connected layer is added to ensure that the output matches the forecasting target dimension.
An overview of the model architecture is sketched in Figure \ref{fig:diagram}.

\section{Experiments on NBA Players Performance Prediction}\label{sec:experiment}
In this section, we go over the experiments we ran to predict NBA players' performances and the assessment of \model{}'s predictive power\footnote{The code and data for this work is available at \url{https://anonymous.4open.science/r/NBA-GNN-prediction}.}. 
We split the dataset (Section \ref{sec:data}) into a training set ($50\%$), validation set ($25\%$), and test set ($25\%$) in chronological order. The input sequence length is 10 and the output sequence length is 1, i.e., we use players' past performance on 10 game days to predict the performance on the next game day.

\subsection{Model Hyperparameters}
The output dimension (\ref{eq:GAT output}) of the GATv2 was set to 32 to avoid overfitting because the input dimension (13) and the final output dimension (6) are both modest. Considering that the player interaction graph is relatively sparse, we chose the number of heads in GAT to be 4. The output dimension of the temporal convolution was set to be 64, which is the same as that in ASTGCN \cite{guo2019attention}. 
Other input settings for the GATv2 layer, including normalized Xavier initialization and dropout, followed the default PyTorch setting. 
We set the dimension of both team embedding and position embedding to 2 for straightforward visualization of the results (Section \ref{subsec:intermediate}).  The model is trained by the Adam optimizer. The learning rate is 0.001 with 0.001 weight decay to avoid overfitting.

\begin{table}
\caption{Performance Comparison with Baseline Models.}\label{table:performance}
\begin{tabularx}{\textwidth}{l | Y Y Y Y}
\hline
 Method & RMSE $\downarrow$ & MAE $\downarrow$ & MAPE $\downarrow$ & CORR $\uparrow$ \\
\hline
N-BEATS & 5.112 & 4.552 & 3.701 & 0.366 \\
 
DeepVAR & 2.896 & 2.151 & 1.754 & 0.396 \\
 
TCN & 2.414 & 1.780 & 0.551 & 0.418 \\
 
ASTGCN & 2.293 & 1.699 & $\mathbf{0.455}$ & 0.453 \\
 
\model{} (Ours) & $\mathbf{2.222}$ & $\mathbf{1.642}$ & 0.513 & $\mathbf{0.508}$ \\
\hline
\end{tabularx}
\end{table}

\subsection{Performance Comparison with Baseline Methods}
We compare the proposed \model{} with the following baseline models. %

\vspace{0.2in}

\noindent 
N-BEATS: a neural network model using double residual stacks of fully connected layers, where each stack represents trend and seasonality respectively \cite{oreshkin2019n}.

\noindent
DeepVAR: a vector autoregressive based probabilistic recurrent neural network (RNN) \cite{salinas2019high}. 

\noindent %
TCN: a variation of convolutional neural network (CNN) for sequence modeling tasks, by combining aspects of RNN and CNN architectures \cite{bai2018empirical}.

\noindent
ASTGCN: a temporal graph convolution network with spatial and temporal attention \cite{guo2019attention}.

\vspace{0.2in}

We use the Python package Darts \cite{herzen2022darts} for N-BEATS, TCN, and DeepVAR implementation. Darts adapts the original N-BEATS architecture to multivariate time series forecasting. 

We use the root mean square error (RMSE), mean absolute error (MAE), mean absolute percentage error (MAPE), and correlation coefficient (CORR) as the evaluation metrics. For CORR, we use a Fisher z transformation \cite{corey1998averaging} to compute the average of correlation coefficients along various output dimension. The results are summarized in Table \ref{table:performance}. We observe that \model{} achieves superior performance on RMSE, MAE, and CORR. The baseline models all capture the dynamic behavior of the various metrics of a player, but most of them fail to account for the interaction among players.

\begin{figure}
\includegraphics[width=1\textwidth]{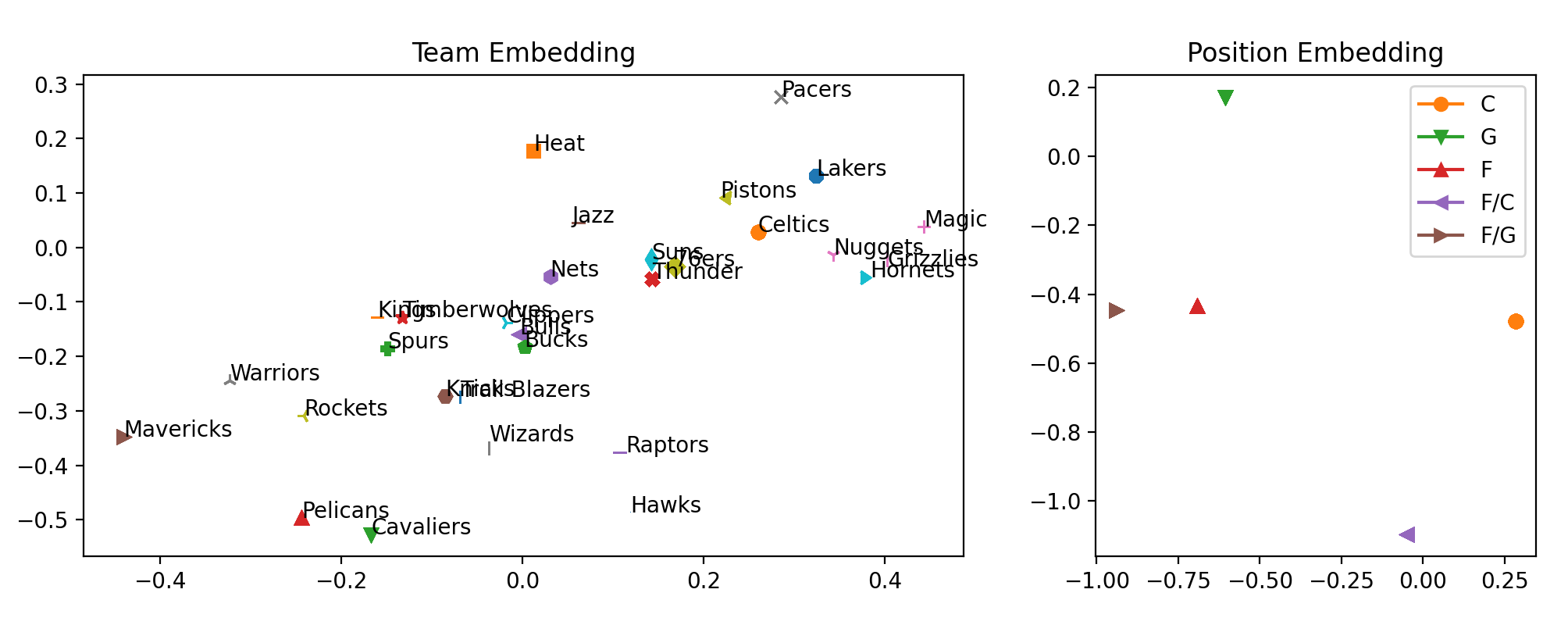}
\caption{The team embedding (left) and position embedding (right) learned using two linear transformation layers in \model{}. Each team is represented by a cluster of points in the 2D team embedding space. Five types of positions are shown in the position embedding. $C$: center, $G$: guard, $F$: forward, $F/C$: forward-center, $F/G$: forward-guard. Note that $F/C$ is not in the middle of $F$ and $C$. Likewise for $F/G$.} \label{fig:embedding}
\end{figure}

\subsection{Interpretation of Model Results}\label{subsec:intermediate}
In this subsection, we look into \model{}'s intermediate results after training. In particular, we examine the team/position embedding and the graph attention. 
\subsubsection{Team Embedding and Position Embedding}
Recall that in Section \ref{subsec:GAT} we use two linear layers to learn the team embedding $team_i$ and player embedding $pos_i$ in \model{}. These embeddings assist the model in accounting for a player's team membership and position when computing his attention to neighboring players. The learned embedding also allows us to visualize the entire team in a low-dimension space. 
In Figure \ref{fig:embedding}, each team is represented by a cluster of points in the 2D team embedding space. Teams with similar play styles and player combinations, such as the Knicks and Trail Blazers, or the Grizzlies and Hornets, are close to each other. Five types of positions are shown in the position embedding. $C$: center, $G$: guard, $F$: forward, $F/C$: forward-center, $F/G$: forward-guard. Note that $F/C$ is not in the middle of $F$ and $C$. Likewise for $F/G$.

\subsubsection{Graph Attention}
\model{} computes player's attention (\ref{eq:attention}) at each time based GATv2. It assigns different weights to players within the same neighborhood, which enables effectively propagating the players' features through the edges of the player interaction graph. %
We demonstrate how players allocate different attention to their teammates and opponents using an example of attention among players from the Phoenix Suns and the Golden State Warriors prior to the game on 2023-01-10. The two teams have played twice (2022-10-25, 2022-11-16) since the 2022-23 season. 

\begin{figure}
\centering
\includegraphics[width=1\textwidth]{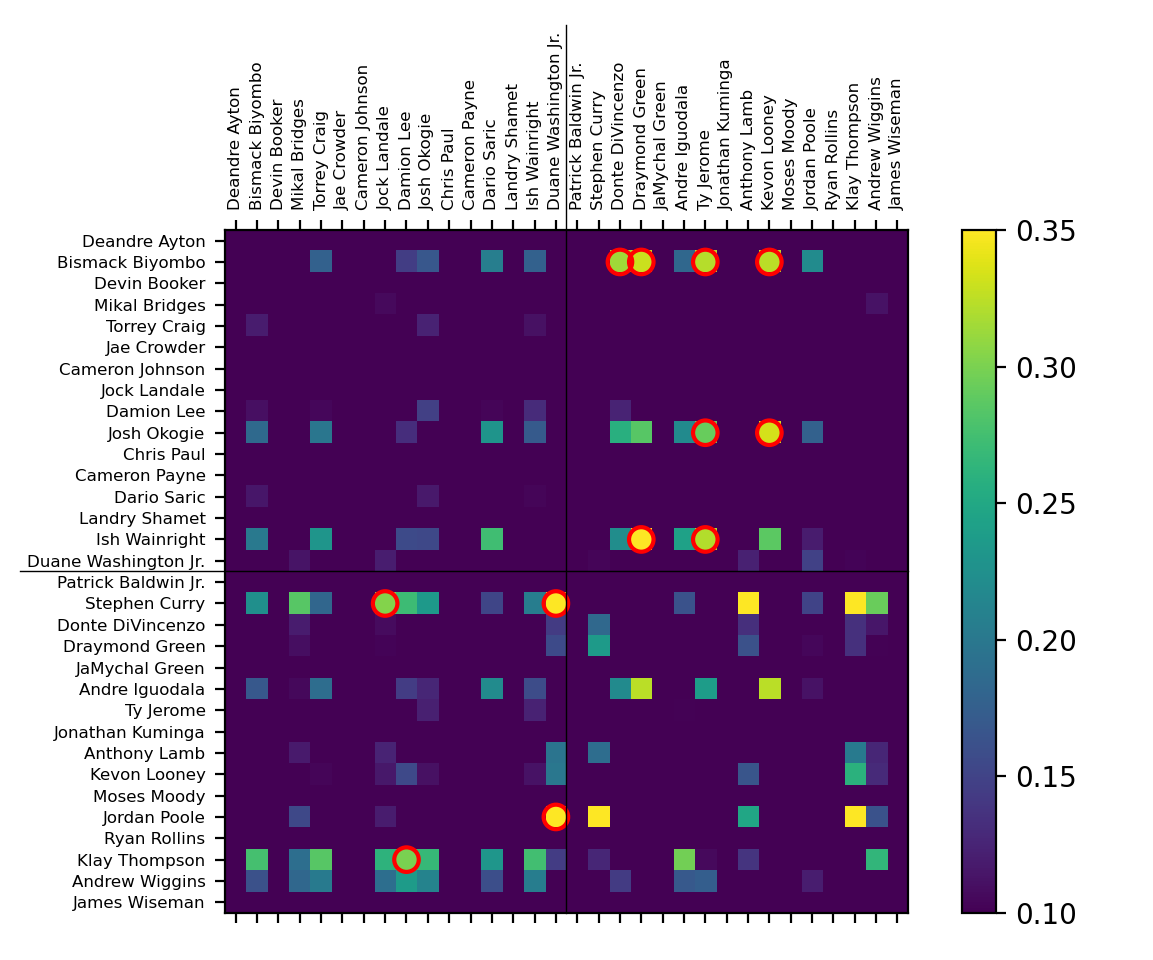}
\caption{The attention heatmap between players from the Suns and the Warriors. Players from the same team are grouped together. The top ten inter-team attention entries are marked with red circles.} \label{fig:attention}
\end{figure}

In the attention heatmap shown in Figure \ref{fig:attention}, each entry $(i, j)$ represents the maximum attention paid to $i$ from $j$ through multiple separate heads. The players are grouped based on their teams. We remove the diagonal entries, which represent the attention one player paid to himself. We also highlight the top 10 entries in the two off-diagonal blocks, i.e., the attention of players between different teams. The Suns' Bismack Biyombo and the Warriors' Stephen Curry received much attention from their opponents. It is also worth noting that the Suns' Damion Lee and the Warriors' Ty Jerome paid much attention to their opponents. Damion Lee was a Warriors player in the 2021-22 season, and Ty Jerome was a Suns player in the 2019-20 season.

\subsection{Case Study: Betting on Underdog Fantasy}\label{subsec:pickem}
\begin{figure}
\centering
\includegraphics[width=1\textwidth]{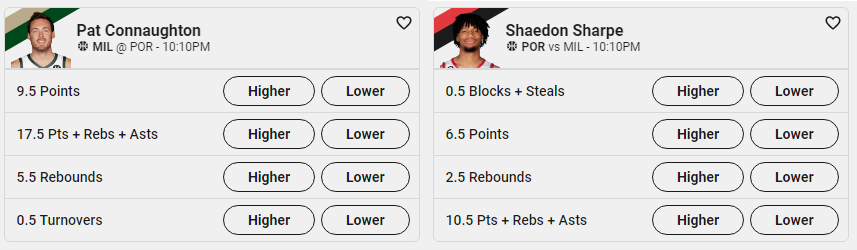}
\caption{The experiment in Section \ref{subsec:pickem} uses the daily Higher-Lower entries from the sports betting website Underdog Fantasy. We use \model{} to predict the player statistics on 2023-01-20.} \label{fig:pickem}
\end{figure}

We conducted an experiment on the sports betting website Underdog Fantasy to demonstrate the practical application of \model{}'s predictive power. An example of the Higher-Lower betting is shown in Figure \ref{fig:pickem}. 

For each NBA game, we use \model{} to predict whether the selected players will achieve certain statistics. We consider it a correct prediction if both the prediction and the true statistic are higher or lower than the value set by the betting website. Using the data from 2023-01-10 to 2023-01-19, the model produced a 35/59 prediction accuracy on 2023-01-20.

\section{Conclusion}
In this study, we tackle the challenge of predicting player performance in sports using a novel deep learning method, \model{}, which leverages graph attention networks to capture the underlying player interactions. To further enhance the model's predictive power, we incorporate a temporal convolution layer to handle multivariate player statistics time series.

Our proposed method has already demonstrated its effectiveness in generating profits in sports betting, while also providing valuable insights for sports analysis and forecasting. However, we believe that further practical improvements could be made by extending the binary-valued adjacency matrix to a weighted adjacency matrix. By encoding additional information, such as the number of minutes two connected players have spent playing together or the number of seasons they have spent together prior to the current one, the model can better capture the nuanced dynamics of player interactions.

\bibliographystyle{splncs04}

\end{document}